\documentclass{article}
\usepackage{spconf,amsmath,graphicx}
\usepackage{amsfonts}
\usepackage{algorithm}
\usepackage{algpseudocode}
\usepackage{array}
\usepackage[hidelinks]{hyperref}
\usepackage{xcolor}
\usepackage{multirow}
\usepackage{makecell}
\usepackage{bbm}

\usepackage{tikz}
\usepackage{pgfplots}
\pgfplotsset{compat=1.6}
\usepackage{caption}

\usepackage{booktabs}

\title{Learning dependencies of discrete speech representations \\ with neural hidden Markov models}
\name{Sung-Lin Yeh, Hao Tang}
\address{School of Informatics, University of Edinburgh}

\begin{document}

\maketitle

\begin{abstract}
While discrete latent variable models have had great success in self-supervised learning, 
most models assume that frames are independent.
Due to the segmental nature of phonemes in speech perception,
modeling dependencies among latent variables at the frame level
can potentially improve the learned representations on phonetic-related tasks.
In this work, we assume Markovian dependencies among latent variables,
and propose to learn speech representations with neural hidden Markov models.
Our general framework allows us to compare to self-supervised
models that assume independence, while keeping the number of parameters fixed.
The added dependencies improve the accessibility of phonetic information,
phonetic segmentation, and the cluster purity of phones,
showcasing the benefit of the assumed dependencies.
\end{abstract}

\begin{keywords}
self-supervised learning, segmental structure, probabilistic graphical models, hidden Markov models
\end{keywords}
\section{Introduction}

Discrete latent variable models for self-supervised learning~\cite{ref:zhou2021comparison,ref:baevski2019vq,ref:baevski2020wav2vec,ref:chung2021w2v} 
have shown strong performance in speech representation learning.
Assuming discreteness in self-supervised models aligns well with phonological categories in speech
perception, and improves the accessibility of phonetic information \cite{ref:baevski2019vq,ref:yeh2022autoregressive}.
Beyond discreteness, dependencies among discrete variables
could be useful not only for speech recognition but also for speech segmentation
and acoustic unit discovery.
However, most self-supervised approaches \cite{ref:baevski2020wav2vec,ref:chung2020vector,ref:yeh2022autoregressive} do not consider dependencies among
discrete latent variables.
In this work, we study the options of learning dependencies
among latent variables in the context of self-supervised learning.

There have been several attempts to incorporate constraints on discrete latent variables.
Slowness penalty \cite{ref:dieleman2021variable}, for example, 
has been applied to encourage slow changes in the encoder outputs.
Similar to slowness, aligned CPC (ACPC) \cite{ref:chorowski21aligned}
enforces piece-wise constant output by aligning future frames to a list of discrete codes.
The act of aligning hints the existence of an underlying probability model.
Inspired by these approaches, we aim to make the probabilistic modeling explicit.

We propose to represent dependencies among discrete variables in the language of probabilistic 
graphical models.
We decide to explore the simplest form of dependecies,
making Markov assumptions on the discrete latent states.
The end result is a neural hidden Markov model (HMM)
in the context of self-supervised learning.\footnote{
Technically, a hidden Markov model is a self-supervised model,
``predicting'' the input through a generative process.}
Neural HMMs are sufficient to achieve the intended effect of ACPC.
To further incorporate slowness, we introduce hops in HMMs.
Similar ideas for modeling slowness in HMMs have been explored in 
\cite{ref:saul1994boltzmann,ref:ghahramani1995factorial}.
In this work, we simply assume multiple chains of low frame-rate HMMs that simultaneously
model dependencies and slowness.

HMMs have been extensively used in modeling transition of symbols,
in particular, in automatic speech recognition \cite{ref:rabiner1989tutorial}.
In contrast, we are interested in the representations learned by
making Markov assumptions, and care less about the learned HMMs.
We empirically compare the representations learned with neural HMMs and with
VQ-APC \cite{ref:chung2020vector} on three probing tasks:
phone classification, phone purity, and phone segmentation.
To isolate the effect of assuming dependencies, we choose to extend the parameterization of VQ-APC to neural HMMs.
VQ-APC becomes a special case of our model if Markov assumptions are removed,
and no additional parameters are needed to parametrize neural HMMs.
We use standard forward-backward algorithm to compute the gradient of HMMs,
allowing us to train our neural HMMs end to end.

In summary, we make the following contributions.
We propose to use neural HMMs to model dependencies among discrete latent variables.
We introduce hops in neural HMMs to model slowness.
The neural HMMs subsume both VQ-APC and ACPC,
and we empirically showcase the benefit of making Markov and slowness assumptions.

\section{Neural Hidden Markov Models}

For a sequence of acoustic frames $x_{1:T}=(x_1, x_2, \dots, x_T)$,
we assume there is a corresponding sequence of discrete latent variables
$z_{1:T}=(z_1, z_2, \dots, z_T)$, for example, a sequence of vector-quantized codes.
The discrete latent variables are typically assumed to be independent,
for example in VQ-APC \cite{ref:chung2020vector}, VQ-CPC \cite{ref:van2020vector}, 
or the family of masked prediction losses \cite{ref:chung2021w2v,ref:baevski2019vq}.
The dependencies are only loosely imposed by the neural networks
that generate the codes.
We focus on imposing Markovian dependencies, leading to neural HMMs.

Given a sequence of frames $x_{1:T}$ and
a sequence of discrete hidden variables $z_{1:T}$,
a neural hidden Markov model defines a joint distribution
\begin{align}
p(x_{k+1:T}, z_{k+1:T}|x_{1:k}) = p(z_{k+1}|x_1)p(x_{k+1}|z_{k+1}) \notag\\
                                 \times \prod_{t=k+2}^{T}p(z_{t}|z_{t-1}, x_{1:t-k}) p(x_{t}|z_{t}),
\label{hmm-fact}
\end{align}
over the frames $x_{1:T}$ and discrete latent variables $z_{1:T}$,
where $k$ is a time shift, similar to the one in APC \cite{ref:chung2019unsupervised} and VQ-APC \cite{ref:chung2020vector}.
The time shift is necessary to avoid degenerate solutions,
and controls what is learned in the representations.
The joint distribution \eqref{hmm-fact} involves a prior distribution $p(z_{k+1}|x_1)$, 
a transition distribution $p(z_{t}|z_{t-1}, x_{1:t-k})$, and an emission distribution
$p(x_{t}|z_{t})$.
These distributions will be parametrized based on neural networks.
As opposed to a regular HMM, our transition distribution from $z_{t-1}$ to $z_t$ depends on
past frames.$x_{1:t-k}$, allowing us to make use of neural networks
to represent $z_{t-1}$ and $z_t$.

\begin{figure}[htp]
\centering\includegraphics[scale=0.8]{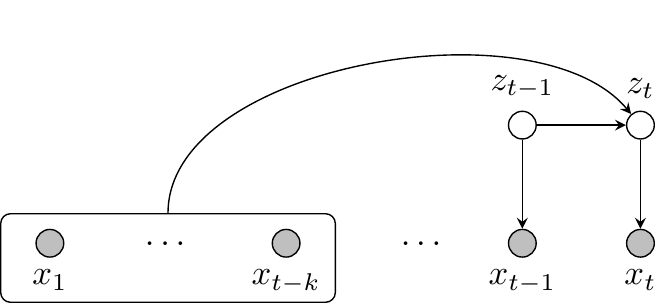}
\caption{The graphical model for the proposed neural HMM.
  Observed variables are shaded.}
\label{fig:hmm}
\end{figure}

\subsection{Neural Parametrization}
\label{sec:hmm-model}

We use a uni-directional recurrent network to compute hidden vectors for each frame.
A linear projection $U$ is used to produce a representation
of a state.
Given a hidden vector $h_{t-k}$, the representation of $z_t$
is defined as
\begin{align}
s_t = h_{t-k}^\top U.
\end{align}

We parametrize the prior distribution as
\begin{align}
    p(z_{k+1}|x_{1}) %
    = \frac{\exp\left(h_1^\top U \mathbf{1}_{z_{k+1}}\right)}{\sum_{j=1}^{N} \exp\left(h_1^\top u_j\right)}
    = \frac{\exp(s_1[z_{k+1}])}{\sum_{j=1}^{N} \exp(s_1[j])},
\end{align}
where $N$ is the number of states, $\mathbf{1}_{z_{k+1}}$ is a one-hot vector with the $z_{k+1}$-th entry
set to 1, $u_j$ is the $j$-th column of $U$, and $s[j]$ is the $j$-th entry of $s$.
In view of vector quantization, the projection $U$ is a codebook for quantizing $h_{t-k}$.

For the transition distribution, 
we compute the representation of a state,
and use the outer product
\begin{align}
    \Phi_t = s_{t-1} s_t^\top = h_{t-k-1}^\top U U^\top h_{t-k},
\label{eq:trans-pot}
\end{align}
to compute the transition probability
\begin{align}
    p(z_{t}| z_{t-1}, x_{1:t-k}) %
    = \frac{\exp(\Phi_t[z_{t-1}, z_{t}])}{\sum_{j=1}^{N} \exp\left(\Phi_t[z_{t-1},j] \right)},
\label{eq:hmm-trans}
\end{align}
where $\Phi_t[i,j]$ is the $ij$ entry of the matrix $\Phi_t$.

Lastly, the emission distribution is defined as
\begin{align}
    p(x_{t}|z_{t}) = \frac{1}{(2\pi)^{d/2}}\exp\left(-\frac{1}{2}\|x_{t} - W V \mathbf{1}_{z_{t}}\|^2\right),
\label{eq:hmm-emission}
\end{align}
where $W$ is a linear projection, $V$ is the codebook, and $d$ is the dimension of $x_{t}$. 
We simply choose $W$ the identity matrix, setting the codeword dimension to $d$.
The choice of Gaussian is aligned with the $L_2$ loss in \cite{ref:yeh2022autoregressive,ref:yang2022autoregressive}.

Note that the parametrization does not introduce any additional parameters
compared to VQ-APC.
If we use a linear form in \eqref{eq:trans-pot}, i.e., removing the edge from $z_{t-1}$ to $z_t$ in Figure \ref{fig:hmm}, then the model falls back
to VQ-APC.
We choose this particular parametrization to isolate
the effect of the Markov assumptions.
Other parametrizations involving more layers are left for future work.

\begin{figure}[htp]
\centering\includegraphics[scale=0.8]{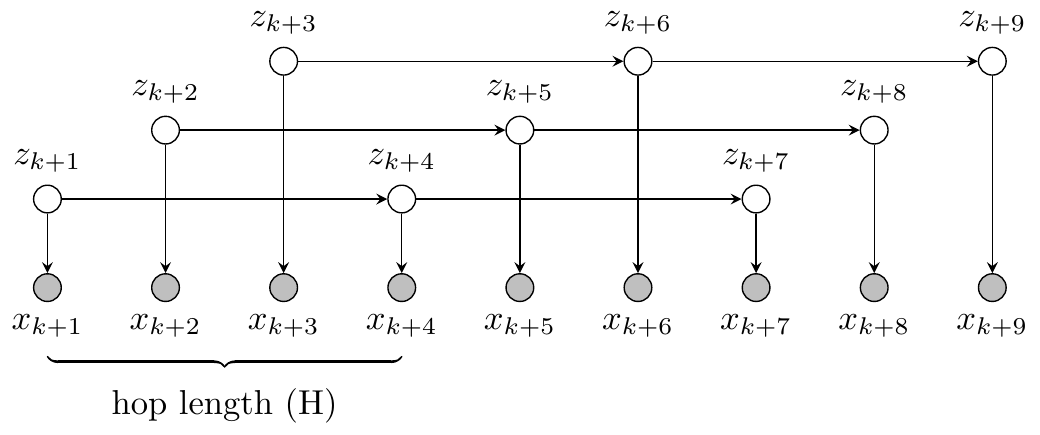}
\caption{An example of low frame-rate HMMs, where we have omitted the edges from the past contexts for 
clarity. The hop length $H$ is $3$ in this example.}
\label{fig:hmms}
\end{figure}

\subsection{Low Frame-Rate Neural HMMs}

The definition of neural HMMs introduces dependencies on every pair of contiguous frames.
However, the probability of staying in a particular state quickly
drops in a few time steps \cite{ref:levinson1986continuously}.
We decide to impose a stronger constraint on the structure of
HMMs to further control slowness.
We introduce hops between consecutive latent variables,
so the models, termed low frame-rate HMMs, are encouraged
to learn dependencies several frames apart.

Low frame-rate HMMs with hops are shown in Figure \ref{fig:hmms}.
Due to hops, a single chain of HMM is not able to cover the generative process
of the entire sequence.
Specifically, if the hop is set to $H$, we use a neural HMM to model
the latent variables every $H$ time steps apart.
To cover every frame in the sequence, we have $H$ independent chains of HMMs,
each of which is shifted by $h$ frames from the first chain, for $h = 1, \dots, H-1$.
The amount of hops to use is a hyperparameter that controls what is learned
in the representations.
We will study how the hop affects the accessibility of phonetic information
in the experiments.

\subsection{Training}

There are several ways to train HMMs.
The de facto approach is expectation maximization \cite{ref:rabiner1989tutorial},
and it is possible to extend this approach to train neural networks
that parametrize the HMMs \cite{ref:tran-etal-2016-unsupervised,ref:ebbers2017hidden}.
We decide to compute the log likelihood using the forward algorithm
and compute the gradients using the standard forward-backward algorithm.
The entire model is trained end to end by maximizing the log likelihood.

The likelihood can be computed by marginalizing $z_{k+1:T}$
with both the forward and the backward algorithm \cite{ref:rabiner1989tutorial}.
Formally, the likelihood is
\begin{align}
   p(x_{k+1:T}|x_{1:k}) &= \sum_{z_{k+1:T}} p(x_{k+1:T}, z_{k+1:T}|x_{1:k}) \notag\\
    &= \sum_{z_{k+1:T}} \prod_{t=k+1}^{T}p(z_{t}|z_{t-1}, x_{1:t-k})p(x_{t}|z_{t}),
\end{align}
where we introduce a sentinel variable $z_k$ to simplify the equations.
The gradient can technically be computed with automatic differentiation,
through the gradient of the forward \cite{ref:eisner2016inside}.
However, we find the approach slow in practice if implemented in Pytorch.
We derive the standard forward-backward algorithm for computing the gradient
of the log likelihood \cite{ref:sutton2012introduction}.
The forward-backward algorithm requires three passes through the sequence,
each taking $\mathcal{O}(TN^2)$ to compute.
We have a custom implementation in C++ to speed up the computation.\footnote{
The implementation of the custom HMM layer will be publically available at \href{https://github.com/30stomercury/hmm-backprop}
{https://github.com/30stomercury/hmm-backprop}.}

\section{Related Work}

Neural HMMs have been used in several unsupervised settings,
for example, structure induction \cite{ref:tran-etal-2016-unsupervised,ref:ebbers2017hidden}.
As we have discussed, the training of HMMs is typically done by
expectation maximization \cite{ref:rabiner1989tutorial}
(or similarly variational inference \cite{ref:beal2003variational}.
Neural HMMs have been used in language modeling and trained
with backpropagation through the forward algorithm \cite{ref:chiu-rush-2020-scaling}.
However, these studies do not concern with learning represenations,
but focus more on the utility of the HMMs for particular tasks.
They also do not involve concepts derived from self-supervised learning,
such as the future-past distinction or masked prediction.

There has also been work on learning structured speech representations,
e.g., aligned CPC (ACPC) and segmental CPC \cite{ref:Bhati2021SegmentalCP,ref:chorowski21aligned}.
Our work is most similar to ACPC \cite{ref:chorowski21aligned},
where dynamic time warping (DTW) was used to align a fixed number of predictions to a short sequence of future frames.
When the number of predictions is fewer than the number of frames, the alignment naturally
imposes a piece-wise constant output.
Our approach technically can subsume ACPC,
but ACPC only aligns the frames to the predictions within a short time span.
In addition, the alignment is done on every time step.
In other words, a frame is used multiple times and aligned to states that are
not the same throughout the sequence.
Our approach does alignment using dynamic programming during training as well, but we ensure that every frame
has a unique latent state assigned.
Subsequent work on ACPC uses a hierarchy to impose slowness \cite{ref:cuervo2022variable},
while we use hops in neural HMMs to achieve a similar effect.
\section{Experiments}

To study the representations learned by our neural HMMs,
we evaluate them on phone classification on Wall Street Journal (WSJ) and phone segmentation
on TIMIT.
We follow the same setting described in prior work 
\cite{ref:chung2020vector,ref:chung2019unsupervised,ref:yeh2022autoregressive}, 
using LibriSpeech \texttt{train-clean-360} %
for pre-training.
Phone classification on WSJ is trained on 90\% of \texttt{si284}, 
and evaluated on \texttt{dev93}.
Following \cite{ref:yeh2022autoregressive}, we also evaluate phone cluster purity on WSJ \texttt{si284}.
For phone segmentation on TIMIT, %
we do not follow the setting in other studies \cite{ref:kreuk2020phoneme,ref:Bhati2021SegmentalCP}.
The published results themselves are not entirely comparable
due to subtle differences, such as data set split and preprocessing protocols.
We instead evaluate segmentation quality on the training set
and favor it over the small test set.
We compute 40-dimensional log Mel features for all data using
a window size of 25 ms and a shift of 10 ms. Features are
normalized with the mean and variance computed from the training sets.

Following \cite{ref:yang2022autoregressive,ref:chung2020vector}, we fix the time shift $k$ to $5$, choosing the recurrent network as 3-layer 512-dimensional unidirectional 
LSTMs.
We will compare training the model with VQ-APC and neural HMMs.
As we have stated, the difference between VQ-APC and neural HMMs
lies in the transition distribution.
The log transition distribution in neural HMMs is quadratic in the state representation.
If the log transition distribution is linear in the state representation,
then the model falls back to VQ-APC.
The two training approaches have the same number of parameters,
and improvements, if any, can be attributed to the added Markov assumptions.
To control all factors including training algorithms,
we choose to marginalize the latent variables in VQ-APC,
instead of using Gumbel softmax \cite{ref:jang2016categorical}.
Incidentally, marginalization leads to an improvement over
Gumbel softmax, as we will see.

For downstream probing tasks,
we freeze the pre-trained models after pre-training on LibriSpeech.
For phone classification,  we train a linear classifier to predict 
phones based on the learned representations.
We use Adam with a learning rate of $10^{-3}$ to train all models.
Pre-trained models are trained for 20 epochs, and the linear classifier is trained for 10 epochs.

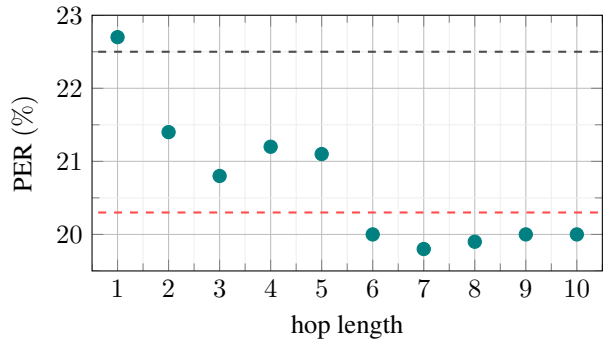
\begin{figure}
\begin{center}
\pgfplotstableread{hop.dat}{\col}
\begin{tikzpicture}
\begin{axis}[
    xmin = 0.5, xmax = 10.5,
    ymin = 19.5, ymax = 23,
    xtick distance = 1,
    ytick distance = 1,
    grid = both,
    minor tick num = 1,
    major grid style = {lightgray},
    minor grid style = {lightgray!25},
    width = 0.47\textwidth,
    height = 0.28\textwidth,
    legend cell align = {left},
    legend pos = outer north east,
    xlabel={hop length},
    ylabel={PER $(\%)$}
]
\addplot[teal, only marks, mark = *, mark size = 2.5pt] table [x = {x}, y = {PER}] {\col};
\addplot[thick, samples=50, smooth,domain=0:6,red!70, dashed] coordinates {(0,20.3)(10.5, 20.3)};
\addplot[thick, samples=50, smooth,domain=0:6,black!70,dashed] coordinates {(0,22.5)(10.5, 22.5)};
\end{axis}
\end{tikzpicture}
\captionof{figure}{Phone classification of neural HMMs with different hops (green points) when $N=100$. The red line indicates the VQ-APC
with marginalization, while the gray line is the VQ-APC with Gumble approximation.}\label{fig:phncls}
\end{center}
\end{figure}

\begin{table}
\begin{center}
\caption{
    A comparison between VQ-APC and neural HMMs on phone classification, phone cluster purity 
    and phone segmentation (\%). 
    Phone segementation are evaluated with a tolerance level of 20 ms using the precision (P), recall (R) and the F1-score.
    The hop length is set to $7$ for neural HMMs.
}
\label{tab:hmm}
\resizebox{\linewidth}{!}{
\begin{tabular}{cccc|ccc}
\toprule
\multicolumn{1}{l}{}                                                                & \multicolumn{1}{c}{N}    & \multicolumn{1}{c}{PER $\downarrow$}    &\multicolumn{1}{c|}{NMI}   &\multicolumn{1}{c}{P} &\multicolumn{1}{c}{R} &\multicolumn{1}{c}{F1}\\
\midrule
\multirow{3}{*}{\begin{tabular}[c]{@{}c@{}}VQ-APC\\ (Marginalization)\end{tabular}}                     & 5      & 23.3             &12.9            &\textbf{42.2}   &67.4            &\textbf{51.9} \\
                                                                                    & 50     & 20.7             &13.4             &24.9            &86.5            &38.7\\
                                                                                    & 100    & \textbf{20.3}    &\textbf{26.4}   &27.6            &\textbf{91.8}   &42.4\\
\midrule                                                                                                                                                              
\multirow{3}{*}{Neural HMM}                                                         & 5      & 22.5             &18.6            &\textbf{56.6}   &79.4            &\textbf{66.1}\\
                                                                                    & 50     & 21.0             &32.5            &31.0            &95.8            &46.6\\
                                                                                    & 100    & \textbf{19.7}    &\textbf{34.6}   &29.4            &\textbf{96.6}   &45.1\\
\bottomrule
\end{tabular}
}
\end{center}
\end{table}

\subsection{Phone Classification}

To simplify the discussion, we use the representations from the second
layer of the LSTMs, as they have consistently shown better results \cite{ref:yeh2022autoregressive}.
Figure \ref{fig:phncls} shows the performance of neural HMMs for different hops with VQ-APC.
We find that VQ-APC using marginalization (red) performs significantly better than using Gumbel approximation (gray)
in \cite{ref:chung2020vector,ref:yeh2022autoregressive}.
We also observe that larger hops (6 frames or larger) perform better than VQ-APC.
The best result happens at a hop of 7 frames, about the average phone duration (80 ms),
which encourages an HMM to capture underlying transitions of phones. 

The PER of various codebook sizes $N$ are reported in Table \ref{tab:hmm},
with the hop length set to 7 frames.
Despite having the same number of parameters, neural HMMs overall performs better VQ-APC.

\begin{figure}
\centering\includegraphics[width=8.5cm]{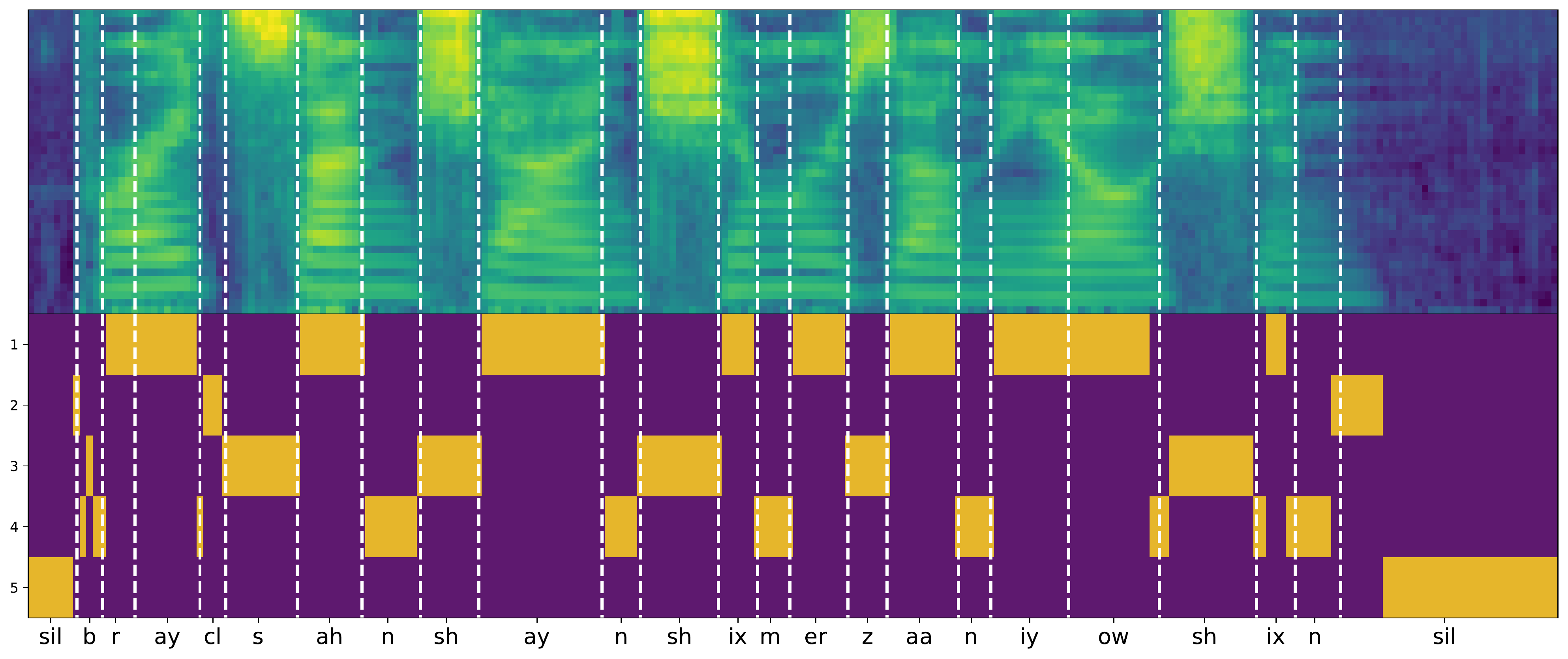}
\caption{An example assignment of frames obtained from the Viterbi algorithm with neural HMMs.
  The top is the log Mel fearure, and the bottom is the decoded states (with $N=5$).
  Ground truth boundaries are shown in white vertical lines.}
\label{fig:align}
\end{figure}

\subsection{Analysis of Learned Codes}

We measure the normalized mutual information (NMI) to evaluate
how well the discrete latent variables match the phonetic categories.
We simply use the Viterbi algorithm to obtain the assignment for each frame.
Results are also in Table \ref{tab:hmm}.
Neural HMMs have a much better normalized mutual information than VQ-APC, suggesting
that the purity of clusters learned by neural HMMs is higher.

For phone segmentation on TIMIT, we simply hypothesize a boundary when a code changes.
Our results do not rely on peak detection \cite{ref:kreuk2020phoneme,ref:Bhati2021SegmentalCP} or any segmentation 
algorithms \cite{ref:kamper2020towards}.
We find that fewer codes leads to fewer transitions, showing good results on phone segmentation.
This is not entirely surprising because in principle we only need to alternate
between two classes if we know the ground truth segmentation.
On the other hand, using more codes leads to oversegmentation,
but reveals finer phonetic information, consistent with the better PER and NMI.
In both cases, neural HMMs greatly improve VQ-APC on phone segmentation. 

Figure \ref{fig:align} is an example of the decoded with a small $N$.
We find the learned codes to correlate with broad phone classes even the model is not trained on TIMIT:
code 1 associates with vowels and semi-vowels; code 2 with closures; code 3 with fricatives;
code 4 with nasals; and code 5 with silence.

\section{Conclusion}

We propose a neural HMM to model the dependencies of discrete latent variables
in the context of self-supervised speech representation learning.
Our neural HMMs generalize both VQ-APC and ACPC.
Under our parametrization, we can isolate the effect of the added Markov assumptions.
Experimental results show that adding dependencies among discrete latent variables
improves the accessbility of phonetic information and the purity of clusters.
Phone segmentation is potentially possible with our model,
though it is less clear whether segmentation can be achieved
by simply using the learned HMMs.
 
\bibliographystyle{IEEEbib}
\bibliography{refs}

\end{document}